\documentclass[conference]{IEEEtran}
\IEEEoverridecommandlockouts
\usepackage{amsmath}
\usepackage{graphicx}
\usepackage{hyperref}
\usepackage{tikz}
 \usepackage{placeins} 
 \usepackage{array}
  \usepackage{caption}
  \usepackage{cite}
  \usepackage{ragged2e}
\usepackage{makecell} 
  \usepackage[caption=false,font=footnotesize]{subfig} 
\usepackage{tikz}

\usetikzlibrary{positioning}   
\usetikzlibrary{arrows.meta}   
\usetikzlibrary{fit}    

 \makeatletter
    \def\ps@IEEEtitlepagestyle{
      \def\@oddfoot{\mycopyrightnotice}
      \def\@evenfoot{}
    }
    \def\mycopyrightnotice{
      {\footnotesize 979-8-3315-7867-1/25/\$31.00~\copyright~2025 IEEE\hfill} 
     \gdef\mycopyrightnotice{}
   }

    \usepackage[mathlines,switch]{lineno}

     \makeatletter
    
    \let\old@ps@IEEEtitlepagestyle\ps@IEEEtitlepagestyle
    \def\confheader#1{%
        \def\ps@IEEEtitlepagestyle{%
            \old@ps@IEEEtitlepagestyle%
            \def\@oddhead{\strut\hfill#1\hfill\strut}%
            \def\@evenhead{\strut\hfill#1\hfill\strut}%
        }%
        \ps@headings%
    }
    \makeatother
    
    \confheader{%
            \parbox{20cm}{2025 28th International Conference on Computer and Information Technology (ICCIT)\\19-21 December, Cox’s Bazar, Bangladesh}
    }

\begin{document}

\title{\LARGE BeHGAN: Bengali Handwritten Word Generation from Plain Text Using Generative Adversarial Networks}

\author{\IEEEauthorblockN{Md. Rakibul Islam \IEEEauthorrefmark{1}, Md. Kamrozzaman Bhuiyan \IEEEauthorrefmark{2}, Safwan Muntasir\IEEEauthorrefmark{3}, Arifur Rahman Jawad\IEEEauthorrefmark{4}, \\Most. Sharmin Sultana Samu\IEEEauthorrefmark{5}}
\IEEEauthorblockA{\IEEEauthorrefmark{1}Department of CSE, Ahsanullah University of Science and Technology, Bangladesh.\\
\IEEEauthorrefmark{2}Enosis Solutions, Bangladesh. 
\IEEEauthorrefmark{3}Smart Technologies BD Ltd. 
\IEEEauthorrefmark{4} Therap BD Ltd.\\
\IEEEauthorrefmark{5}Department of CSE, BRAC University, Bangladesh.}
\IEEEauthorblockA{Email: \IEEEauthorrefmark{1}rakib.aust41@gmail.com,  \IEEEauthorrefmark{2}kamrozzamaan@gmail.com, \IEEEauthorrefmark{3}safwanm.cse@gmail.com, \IEEEauthorrefmark{4}arjawad.cse@gmail.com, \\ \IEEEauthorrefmark{5}sharminsamu130@gmail.com
}
}

\twocolumn[
\begin{@twocolumnfalse}
\vfill

\fontsize{20}{24}\selectfont

\textbf{IEEE Copyright Notice}

\vspace{1em}

\fontsize{14}{17}\selectfont   

\noindent
\begin{minipage}{1.0\textwidth}
\justifying

\textcopyright\ 2025 IEEE. Personal use of this material is permitted. Permission from IEEE must be obtained for all other uses, in any current or future media, including reprinting/republishing this material for advertising or promotional purposes, creating new collective works, for resale or redistribution to servers or lists, or reuse of any copyrighted component of this work in other works. \\

\vspace{2em}

This work has been accepted for publication in \textbf{2025 28th International Conference on Computer and Information Technology (ICCIT)}. The final published version will be available via IEEE Xplore. \\

DOI: \textit{TBD}

\end{minipage}

\vfill
\end{@twocolumnfalse}
]

\maketitle
\begin{abstract}
Handwritten Text Recognition (HTR) is a well-established research area. In contrast, Handwritten Text Generation (HTG) is an emerging field with significant potential. This task is challenging due to the variation in individual handwriting styles. A large and diverse dataset is required to generate realistic handwritten text. However, such datasets are difficult to collect and are not readily available. Bengali is the fifth most spoken language in the world. While several studies exist for languages such as English and Arabic, Bengali handwritten text generation has received little attention. To address this gap, we propose a method for generating Bengali handwritten words. We developed and used a self-collected dataset of Bengali handwriting samples. The dataset includes contributions from approximately five hundred individuals across different ages and genders. All images were pre-processed to ensure consistency and quality. Our approach demonstrates the ability to produce diverse handwritten outputs from input plain text. We believe this work contributes to the advancement of Bengali handwriting generation and can support further research in this area.\\

\renewcommand
\IEEEkeywordsname{Keywords}


\end{abstract}

\section{Introduction}

Handwritten text has been a fundamental medium for recording knowledge, culture and official documentation. Despite the widespread adoption of digital technologies, handwritten text remains important in various sectors. Healthcare, education and financial institutions continue to rely heavily on handwritten documents for communication, record-keeping and verification \cite{tabassum2022online}.

In recent years, Handwritten Text Recognition (HTR) has gained significant research attention. However, Handwritten Text Generation (HTG), the task of producing synthetic handwritten images from typed text, is still emerging as a distinct and promising area \cite{bhunia2021handwriting, fogel2020scrabblegan, kang2020ganwriting, pippi2023handwritten}. HTG has the potential to support applications in many areas. It can create synthetic data for training HTR models and help solve data shortage problems. It can assist people with physical disabilities by converting typed text into handwritten form. It can support design tasks by producing diverse handwriting styles. HTG can also help in font creation, restoration of historical documents and handwriting analysis. It can rewrite or correct handwritten documents automatically.

HTG can be approached in two ways: online and offline. The online method \cite{aksan2018deepwriting, ji2019generative, kotani2020generating} focuses on reproducing the dynamic trajectory of handwriting using pen stroke data. However, collecting such data is expensive and impractical for many use cases, especially historical document synthesis. The offline method \cite{bhunia2021handwriting, kang2020ganwriting, luo2022slogan, fogel2020scrabblegan}, which generates static handwritten images, is more feasible and has become the preferred approach in recent works. These systems learn from image-based handwriting datasets and use style encoders to capture the visual features of different writing styles.

Several studies have been conducted for English and Arabic using public datasets such as IAM \cite{marti2002iam}, RIMES \cite{grosicki2009icdar}, CVL \cite{kleber2013cvl} and AHCD \cite{el2017arabic}. However, there is no significant work in generating Bengali handwriting, particularly at the word level. To the best of our knowledge, Bengali HTG research is limited to digit generation and no standard Bengali word-level handwriting generation dataset exists. This gap presents a clear research question, \textbf{“Can we develop a GAN-based system that generates word-level Bengali handwritten text in diverse human writing styles from typed input?”} 

\begin{figure}[hbt!] 
    \centering
    \includegraphics[width=90mm,scale=0.6]{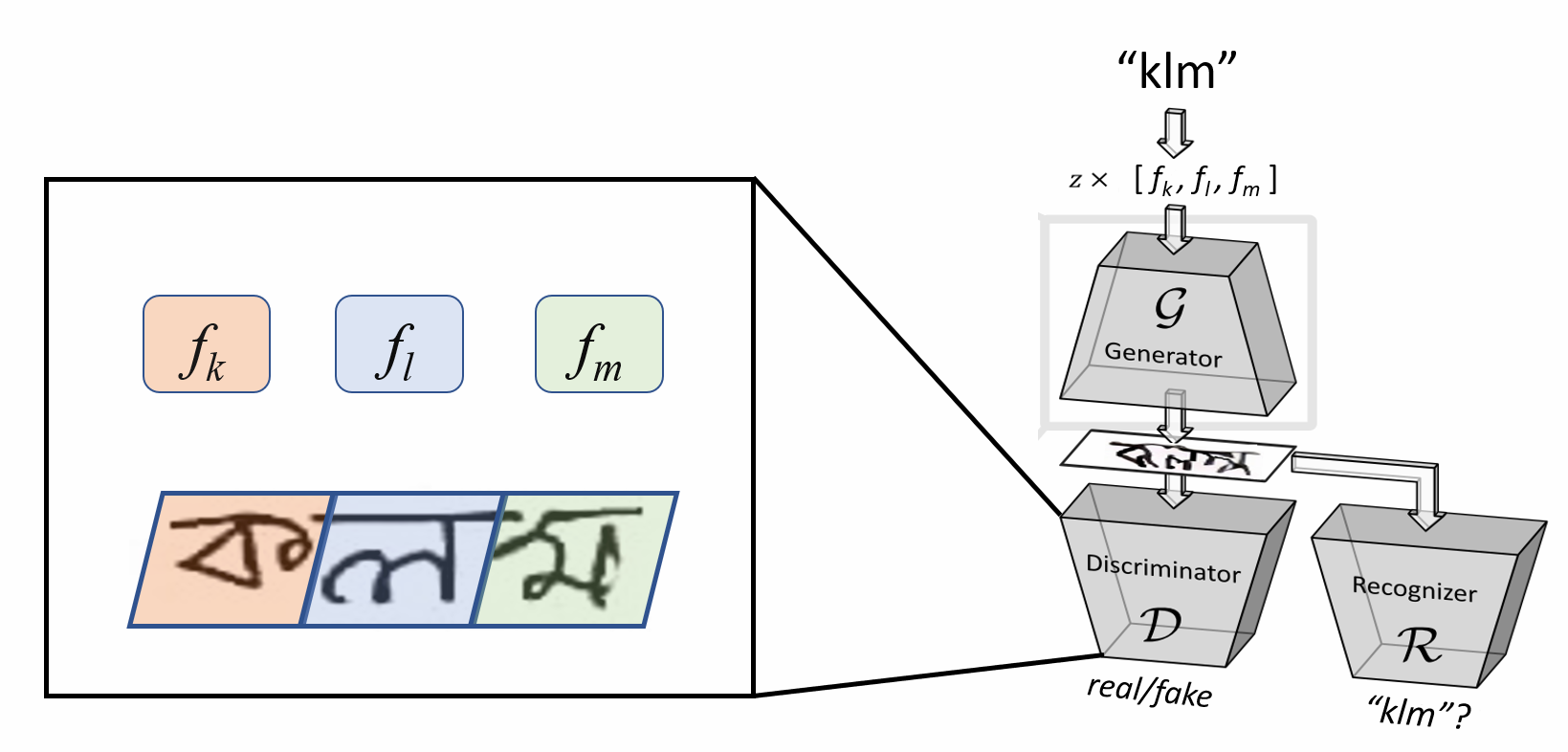}
    \caption{Word Generation Architecture (inspired by \cite{fogel2020scrabblegan}). The input Bengali word is mapped to the English letters ``\textbf{klm}". The model selects the corresponding character samples from the dataset and generates a mask for each character. Due to the overlapping receptive field property of CNNs, the generator produces each character individually. The generated characters are then combined with overlapping regions to form the final handwritten word.}
    \label{fig:gr6}
\end{figure}

In this work, we aim to answer this question by introducing BeHGAN, a GAN-based framework for Bengali handwritten word generation. We approach this problem using the offline HTG paradigm. Our system takes plain Bengali text as input and produces stylized handwritten word images. To support this task, we created a custom dataset by collecting handwriting samples from around five hundred individuals across different age groups, genders and occupations. This dataset captures diverse writing styles and serves both training and evaluation purposes.

To summarize, our contributions include:
\begin{itemize}
    \item Proposing BeHGAN, a GAN-based model for generating Bengali handwritten word images from typed input.
    \item Creating a new Bengali handwriting dataset suitable for generation tasks with samples from five hundred participants.
    \item Demonstrating the effectiveness of our approach in generating realistic handwritten images and opening new directions for Bengali HTG research.
\end{itemize}

The structure of this paper is as follows. Section II reviews existing research and highlights related work. Section III outlines the background study relevant to this research. Section IV describes the methodology, including the technical process and the strategy used for generating handwritten words. Section V presents the experimental results and offers a discussion. Finally, Section VI provides the conclusion and suggests directions for future research.

\section{Related Works}
Several studies have explored the task of handwritten text generation using different architectures and datasets. \cite{fogel2020scrabblegan} proposed a semi-supervised method that integrates a discriminator and a text recognition network to ensure both visual realism and text accuracy in the generated images. The system uses a noise vector to maintain writing style across characters. It performs well on RIMES \cite{grosicki2009icdar}, IAM \cite{marti2002iam} and CVL \cite{kleber2013cvl} datasets, though it is limited to fixed image dimensions and shorter words. In \cite{fahim2019bangla}, a semi-supervised GAN was used to generate Bengali handwritten digits using the BHAND \cite{rahman2016towards} dataset, applying a multi-layer convolutional architecture with dropout and softmax layers. While effective for digit generation, the model is not extended to word-level generation. JokerGAN \cite{zdenek2021jokergan} also uses IAM \cite{marti2002iam} and CVL \cite{kleber2013cvl} datasets, relying on variations of a latent vector to control writing style. However, it focuses mainly on English text.

In contrast, \cite{mustapha2022conditional} introduced CDCGAN for Arabic character generation with a multiclass discriminator to classify both real and generated characters. Though innovative in classification design, the model faces instability issues in training and limitations in dataset diversity. \cite{nishat2019synthetic} used a conditional GAN with an encoder-decoder structure to generate Bengali characters from datasets like BanglaLekha-Isolated \cite{biswas2017banglalekha} and CMATERDB. The approach excludes word-level generation, which is mentioned as a future goal. \cite{ghosh2017handwriting} applied a modified DCGAN combined with reinforcement learning to improve generation quality and spacing for handwritten alphabets. Although promising for applications like signature verification and document digitization, their model focuses on character-level generation and only forms basic words later through post-processing.

\cite{vanherle2024vatr++} focuses on input preparation and training regularization using GANs, Transformers and Diffusion models to improve HTG model performance and generalization. Its main contribution is a standardized evaluation protocol to resolve benchmarking inconsistencies, though further refinement in training and evaluation is needed. \cite{pippi2023handwritten} introduces a Transformer-based few-shot handwritten text generation model that uses visual archetypes from GNU Unifont glyph images to represent rare characters and unseen styles. Pre-training on a large synthetic dataset improves generalization, but the model struggles with fine style details, highlighting the need for better stylistic modeling and data diversity. \cite{zdenek2023handwritten} presents a vision transformer-based style encoder that creates character-specific encodings from reference images, enabling high-fidelity style imitation and random style generation through latent vector sampling. It generalizes well to unseen styles and words, setting a new benchmark in handwriting imitation.

Across all studies, most models perform well for individual characters or digits, but few address full word generation in low-resource languages like Bengali. Limitations such as fixed image sizes, lack of stylistic diversity or unstable training persist, leaving room for future improvements in multi-style, word-level handwritten text generation. 

\section{Background Study}
\subsection{Semi-Supervised Generative Adversarial Network (SGAN)}
Semi-Supervised Generative Adversarial Network (SGAN) \cite{odena2016semi} extends the standard GAN for semi-supervised learning. It learns from both labeled and unlabeled data that improves performance with limited labeled samples. The discriminator operates in two modes: unsupervised, where it distinguishes real from fake data, and supervised, where it predicts class labels. It is trained to predict \( C+1 \) classes, with \( C \) for real classes and \( 1 \) for fake data. This setup enables feature learning from unlabeled data and classification of labeled data. The discriminator can also be used for classification through transfer learning. It helps improve performance when there are fewer labeled examples.
\subsection{Generative Facial Prior-GAN}
GFP-GAN \cite{wang2021towards} is a face restoration model that combines a generative adversarial network with a facial prior to enhance degraded images. It restores high-resolution faces by reducing noise and preserving identity. Unlike traditional inversion methods, it uses a pre-trained face GAN and a degradation removal module to recover features. The model extracts latent features, converts them into style vectors with cross-layer perceptrons and applies spatial features to refine the output. Unchanged features are preserved. It uses perceptual reconstruction and adversarial loss for training. GFP-GAN produces high-quality restorations while balancing realism and fidelity.
\subsection{Evaluation Metrics}\label{A9}
The Structural Similarity Index Measure (SSIM) \cite{wang2004image} measures visual similarity between generated and real images using luminance, contrast and structure. Higher SSIM indicates better quality. The Fréchet Inception Distance (FID) \cite{heusel2017gans} compares feature distributions of real and generated images with lower scores showing better quality and diversity. The Geometric Score \cite{khrulkov2018geometry} evaluates handwritten text generation by measuring structural similarity of generated and real word images. It focuses on stroke alignment, spacing, height and writing flow. Lower scores indicate closer resemblance to real handwriting.

\section{Methodology}
\subsection{Model Architecture}
\begin{figure}[hbt!] 
    \includegraphics[width=85mm,scale=0.7]{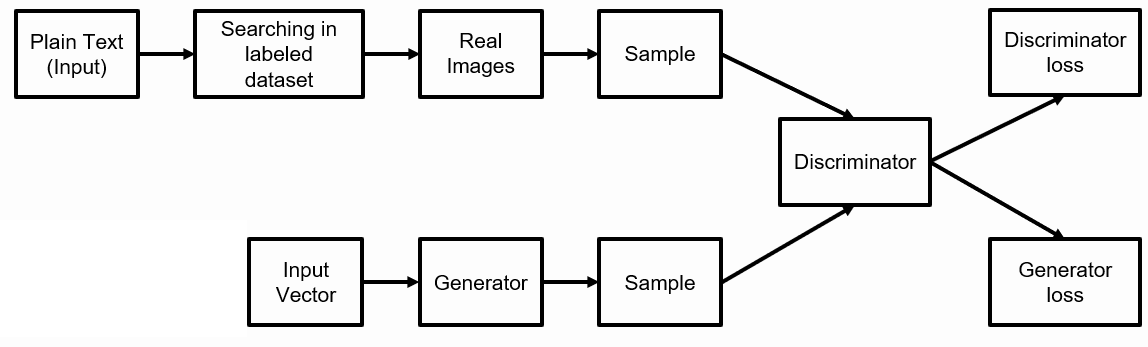}
    \caption{Proposed Methodology}
    \label{fig:gr5}
\end{figure}

We used a semi-supervised GAN for handwriting generation with labeled data. The model was trained to generate Bengali handwritten words from plain text input. The generator learned handwriting styles from the dataset and produced realistic samples. The discriminator evaluated these samples and provided feedback. The generator improved by attempting to fool the discriminator, while the discriminator refined its ability to detect differences. This adversarial process continued until the generator produced realistic handwritten text. Figure \ref{fig:gr5} illustrates the proposed methodology.

The discriminator focused only on evaluating the generated handwriting and giving feedback. The generator had a dual role. It learned handwriting styles from the dataset and attempted to fool the discriminator by producing human-like samples. The discriminator provided an error rate as feedback and the generator tried to minimize this error to improve the handwriting quality. The architecture combined the discriminator \( D \) and a text recognition network \( R \). The network \( R \) ensured that the generated handwriting matched the input text, while \( D \) promoted realistic writing styles. The final loss function was defined as \( L = L_D + \gamma L_R \), where \( L_D \) and \( L_R \) are the loss terms for the discriminator and recognition network. The main novelty of our system was in the design and training of the generator.

The design is based on the observation that handwriting is a regional process where each letter is influenced only by its immediate predecessor and successor. Instead of generating a full word as a single image, the generator \( G \) produces each character individually while using the overlapping receptive fields of CNNs to account for neighboring letters. \( G \) is composed of identical character-based class-conditional generators, each generating a mask for its input character. Convolutional upsampling layers enlarge the receptive fields, which increases overlap between adjacent characters and enables smooth transitions. By learning contextual relationships, the network generates different variations of the same character depending on its position. The handwriting style is controlled by a noise vector \( Z \) that remains constant for all characters of a word to maintain uniformity.

The discriminator \( D \) distinguishes synthetic images from real ones and also evaluates handwriting style. It uses a convolutional structure that can process images of varying sizes and contains multiple "real/fake" classifiers with contiguous receptive fields. Character-level annotations are not used, so no class supervision is applied, allowing \( D \) to be trained with unlabeled images from external datasets. Outputs of all classifiers are combined using a pooling layer to generate the final discriminator score. The recognizer \( R \) checks if the generated text is meaningful by comparing recognized text with the input text and penalizing \( G \) for mismatches. \( R \) is trained only on real annotated handwritten samples. The generator is optimized using the adversarial loss \( L_D \) and the recognizer loss \( L_R \). The relative weight of \( L_R \) with respect to \( L_D \) is controlled by the parameter \( \gamma \): \( L_G = L_D + \gamma L_R \). 

\subsection{Generation Workflow}
Figure \ref{fig:gr6} illustrates the architecture for generating a specific word. The input word is first provided as plain text by mapping it to English letters. The model identifies each character in the input word and retrieves the corresponding samples from the dataset. For every character, it generates a mask that is multiplied with a noise vector \( Z \). This masked input is then passed to the generator, which produces the handwritten sample of the word. The generated image is sent to both the discriminator \( D \) and the recognizer \( R \). The discriminator evaluates whether the image is real or synthetic, while the recognizer checks if the generated word matches the input text.

\subsection{Experimental Setup}
We reproduced the code from \cite{fogel2020scrabblegan} using our dataset. The original code was developed for an English handwritten image dataset, while our dataset contained Bengali handwritten images. As a result, the original code was not directly compatible with our data. We modified the code to make it functional with our dataset. We used Google Colaboratory as the code editor platform. The default GPU was used to accelerate the output generation process. We utilized various Python libraries such as Pandas, NumPy and PyTorch to implement the work.

\subsection{Dataset Collection and Preprocessing}
We did not find any previous work on word-level generation for Bengali handwritten text. Therefore, we designed an experiment with a limited set of alphabets and words. We could not find any suitable dataset for this task. Hence, we created our own dataset. We selected five alphabets from the fifty alphabets of the Bengali language. We formed words by combining these alphabets. We created two-character words by combining any two of the five alphabets and three-character words in the same way. In total, we prepared 30 words. These included 5 single-character words, 16 two-character words and 9 three-character words. We wrote these words on a white page to prepare the data collection sheet.

\begin{figure}[htbp]
    \includegraphics[width=70mm,scale=0.50]{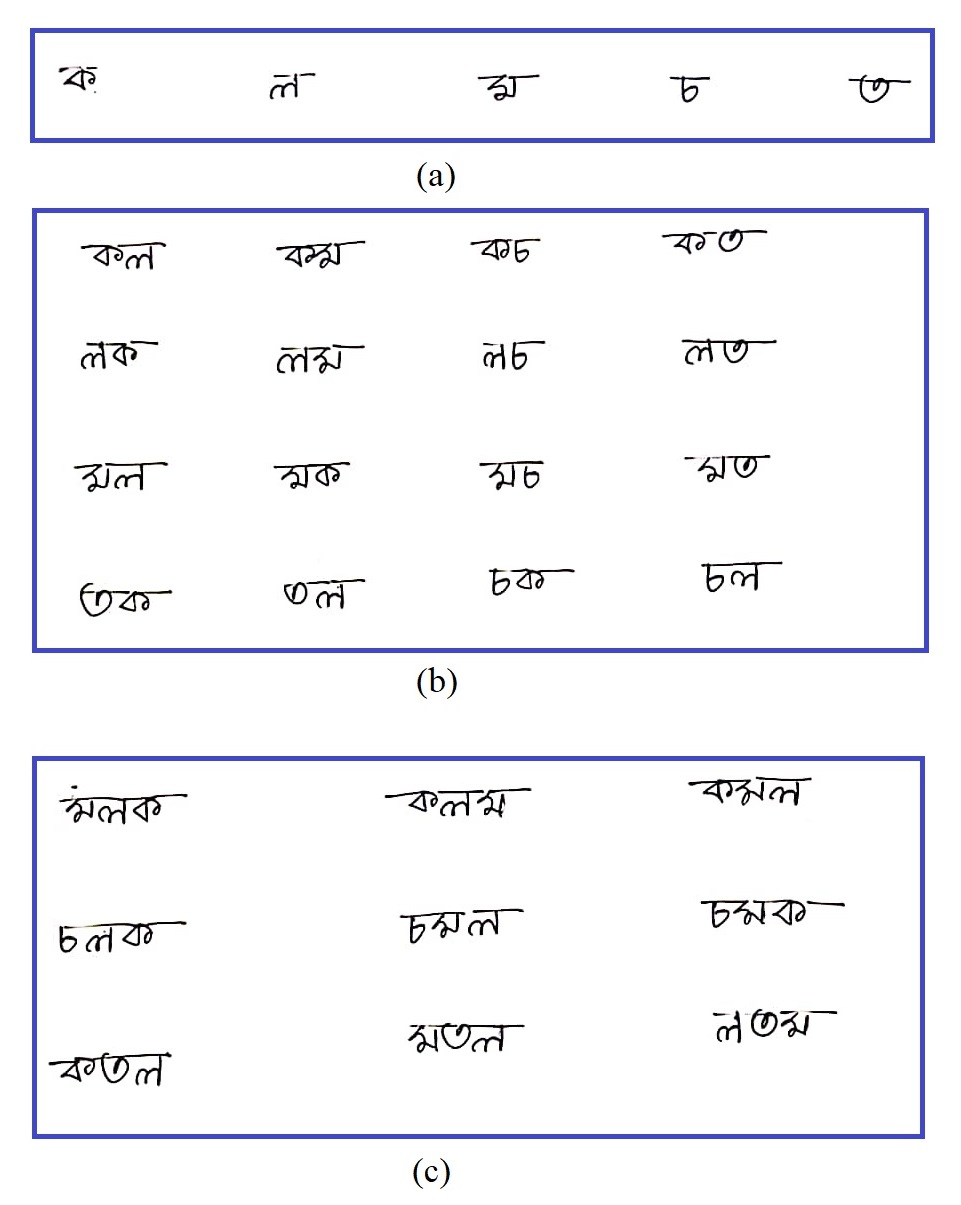}
    \caption{Data Collection Format}
    \label{fig:gr1}
\end{figure}

We collected handwriting samples from approximately five hundred participants of different ages, genders and professions through both online and offline methods. The process required a long time because each participant had to write all 30 words. We approached around seven hundred individuals for this task. Many volunteers participated eagerly after understanding the purpose of our work. However, some individuals refused to participate after seeing the writing sheet. Figure \ref{fig:gr1} presents the data collection format used in this study. Part (a) represents the single-character words. Part (b) represents the two-character words. Part (c) represents the three-character words.

\begin{figure}[htbp]
    \centering
    \includegraphics[width=70mm,scale=0.4]{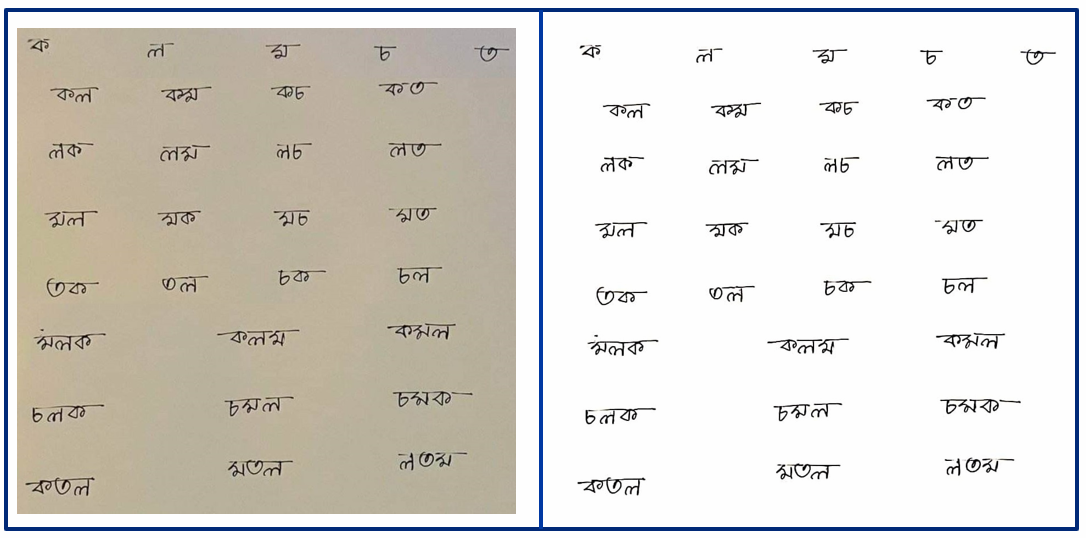}
    \caption{Before and After applying filter}
    \label{fig:gr2}
\end{figure}

After collecting handwriting samples from five hundred individuals, we performed preprocessing to construct the dataset. We used a black pen on white paper to keep all samples uniform. As we captured photos of the handwritten pages, the images varied in brightness and contrast. Their backgrounds also appeared different. To address this, we applied a filter to make the background of each image consistent in brightness and contrast. Figure \ref{fig:gr2} shows the raw data and the corresponding filtered data. The filtered images have a white background.

\begin{figure}[hbt!] 
    \centering
    \includegraphics[width=70mm,scale=0.4]{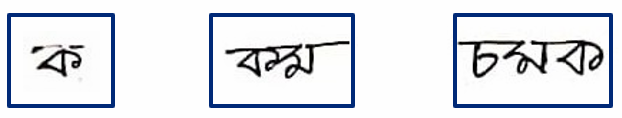}
    \caption{Individual word samples after cropping}
    \label{fig:gr3}
\end{figure}

We selected only the well-written images and discarded the rest. The dataset included handwriting from people of different ages and education levels, including school children. Some samples were unclear or had poor image quality. We removed those samples to maintain the quality of the dataset. After selection, we cropped each word from the images and stored them in separate directories based on word length. Figure \ref{fig:gr3} shows examples of single-character, two-character and three-character word images after cropping. The image sizes remained random at this stage and were fixed in the next preprocessing step.

\begin{figure}[hbt!] 
    \centering
    \includegraphics[width=80mm,scale=0.5]{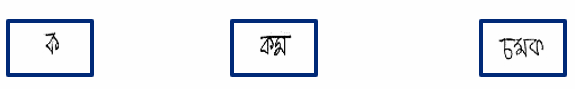}
    \caption{Individual word samples after preprocessing}
    \label{fig:gr4}
\end{figure}

We resized the images in the next preprocessing step. We set the width and height to 16×32 pixels for single-character words, 32×32 pixels for two-character words and 48×32 pixels for three-character words. To match the common style of writing with a black pen on white paper, we converted the images to grayscale. We set the bit depth to 8, which produced grayscale images with a white background. Figure \ref{fig:gr4} shows the samples of single-character, two-character and three-character word images after preprocessing. The resolution of the images decreased due to the small image size. We used a GAN model to generate handwriting, which required a large number of images. After completing all preprocessing steps, we had a total of ten thousand images. To further increase the dataset size, we performed data augmentation.

\subsection{Dataset Organization}
We organized the dataset into three categories: single-character words, two-character words and three-character words. For single-character words, we collected handwritten images of five selected alphabets shown in Figure \ref{fig:gr1}. For two-character and three-character words, we collected handwritten images of combinations of these five alphabets as shown in Figure \ref{fig:gr1}. We stored the images of each category in separate directories to maintain proper organization.

Since we used a semi-supervised GAN, labeling of images was necessary. We created text files for labeling, where each image had a corresponding text file with the same name. After preprocessing, the dataset contained 2,150 single-character words, 4,850 two-character words and 3,000 three-character words. We then applied data augmentation and increased the dataset size to four times the original count.

\begin{table}[!htp]
\normalsize
\centering
\caption{\label{tab:dataset} Total samples in our dataset. One, Two and Three means Single-character, Two-character and Three-character words respectively.}
\begin{tabular}{|c|c|c|c|c|c|}
\hline
Dataset Type & One & Two & Three  \\ 
\hline
Raw & 2150 & 4850 & 3000\\ 
\hline
Augmented & 8600 & 19400 & 12000 \\ 
\hline
\end{tabular}
\end{table}

\section{Result Analysis}
\begin{figure}[hbt!] 
    \centering
    \includegraphics[width=70mm,scale=0.40]{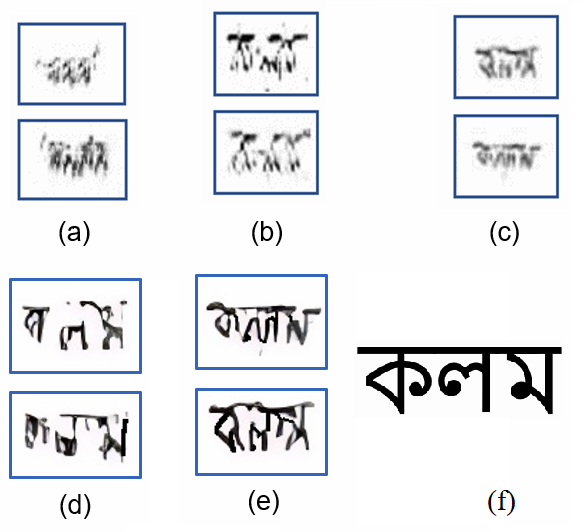}
    \caption{Generated output samples (a-e) of different stages of our experiments for the plain text input (f). }
    \label{fig:gr7}
\end{figure}

In this section, we present the experimental results from the initial stage to the final outcome. The experiments are described using the example of generating a three-character Bengali word from input in plain text format, as shown in part (f) of Figure \ref{fig:gr7}.

In Figure \ref{fig:gr7} (a), the generated output is very noisy because the model was trained on a small dataset containing only 2,756 handwritten images. GANs require a large dataset, so the limited data resulted in poor quality output. In Figure \ref{fig:gr7} (b), the output improved slightly after oversampling the real images to increase the dataset size. After oversampling, the dataset contained 24,804 handwritten images. Although the output was less noisy than in  Figure \ref{fig:gr7} (a), the quality was still unsatisfactory. In Figure \ref{fig:gr7} (c), the generated word became clearer than in Figure \ref{fig:gr7} (a) and Figure \ref{fig:gr7} (b), but the resolution remained low. To improve the results, we applied data augmentation techniques such as scaling, rotation and translation to every original image, which increased the dataset to 27,532 handwritten images.

In Figure \ref{fig:gr7} (d), the structure of each character became more recognizable and the word was readable, but the image was still hazy. To solve this, we collected additional handwriting samples from new participants, performed preprocessing and added them to the dataset. We also used GFP-GAN, which helped overcome the resolution issue observed in Figure \ref{fig:gr7} (c). In Figure \ref{fig:gr7} (e), the output became much clearer and the words were easily readable. This improvement was achieved by using GFP-GAN along with a larger dataset containing five alphabets and their combinations, as shown in Figure \ref{fig:gr1}. The increased variation in words contributed to the better quality compared to the previous outputs.

The training process was monitored across multiple epochs to identify the point at which the model produced the most visually accurate and structurally consistent handwritten words. At each epoch, generated outputs were analyzed for clarity, character alignment and stroke continuity. Early epochs often produced noisy and distorted characters, while intermediate epochs demonstrated gradual improvements in structure and readability. Prolonged training beyond a certain stage led to minor gains or slight degradation due to overfitting. An optimal number of epochs was selected based on both quantitative metrics and visual inspection. SSIM, FID and geometric scores were calculated at each checkpoint. The epoch yielding the highest SSIM, lowest FID and the best geometric consistency was considered optimal.

\begin{table}[!htp]
\normalsize
\centering
\caption{\label{tab:evaluation} Evaluation results of the generated Bengali handwritten image for the final outcome.}
\begin{tabular}{|c|c|}
\hline
\textbf{Evaluation Metric} & \textbf{Score} \\ 
\hline
SSIM & 0.67 \\ 
\hline
FID & 41 \\ 
\hline
Geometric Score & 0.63 \\ 
\hline
\end{tabular}
\end{table}

Table \ref{tab:evaluation} presents the evaluation results for the generated Bengali handwritten images as the final outcome. SSIM measures structural similarity between generated and real images. SSIM values range from 0 to 1, where values closer to 1 indicate higher similarity. Our model achieved an SSIM of 0.67, which indicates moderate similarity. FID evaluates the quality of generated images compared to real images. Lower FID scores indicate better quality and scores below 50 are generally acceptable for handwriting generation tasks. Our model achieved an FID of 41, which shows reasonable image quality. The geometric score measures the structural correctness of character shapes. Its values range from 0 to 1, where higher values indicate better geometry. The obtained geometric score of 0.63 shows that the generated characters maintain acceptable structural accuracy.\\ \\
\textbf{Ablation Study}\\
An ablation study was performed to evaluate the impact of each major component and training configuration on the final performance. Different variations of the model were trained by systematically removing or modifying individual elements such as data augmentation, dataset size and post-processing enhancement. Evaluation was based on SSIM, FID and geometric score.

\begin{table}[!htp]
\normalsize
\centering
\caption{\label{tab:ablation} Ablation study results. GS is used as the abbreviation of Geometric Score.}
\begin{tabular}{|l|c|c|c|}
\hline
\textbf{Configuration} & \textbf{SSIM} & \textbf{FID} & \textbf{GS} \\
\hline
Baseline (limited dataset) & 0.42 & 78 & 0.37 \\
\hline

Oversampling & 0.51 & 66 & 0.44 \\
\hline

Data Augmentation & 0.58 & 54 & 0.51 \\
\hline

Additional Samples & 0.62 & 46 & 0.58 \\
\hline

GFP-GAN Enhancement & \textbf{0.67} & \textbf{41} & \textbf{0.63} \\
\hline
\end{tabular}
\end{table}

The results show a consistent improvement in image quality and structural accuracy as each enhancement was added. Dataset expansion and augmentation significantly reduced noise and improved character clarity. The use of GFP-GAN provided the final boost, producing high-resolution outputs with better geometric consistency. This demonstrates that both dataset diversity and post-processing refinement play key roles in achieving optimal performance.

\section{Conclusion and Future Works}
We introduced an initial approach to Bengali handwritten word generation using a semi-supervised GAN model. A custom dataset was created by collecting handwriting samples from five hundred individuals and applying augmentation due to the lack of public dataset. The model was trained on five Bengali alphabets mapped to English letters to generate single words. The system has limitations, including limited alphabet coverage, no support for conjunct characters and low output resolution. Future work will expand the dataset to include all Bengali alphabets, enable direct recognition of Bengali characters and enhance quality using advanced GAN and diffusion variants. The long-term goal is to generate any Bengali handwritten word or sentence from text input. Our dataset can also support handwritten text recognition and related research.

\bibliographystyle{ieeetr} 

\addcontentsline{toc}{chapter}{References}
\bibliography{ref}

\end{document}